\documentclass[12pt]{article} 
\usepackage{graphicx,latexsym}
\newcommand{\bi}[1]{\mbox{\boldmath $#1$}}

\newtheorem{proposition}{Proposition}

\newcommand{\mylag}{\alpha}
\newcommand{\mycal}{\mathcal}
\newcommand{\inner}[2]{#1\cdot#2}
\newcommand{\wt}{\omega}

\newcommand{\myft}[1]{{#1}^*}
\newcommand{\myftx}{\myft{\bix}}

\newcommand{\mycd}[1]{\hat{#1}}
\newcommand{\transpose}{^{\mathsf T}}
\newcommand{\zeroth}{^{(0)}}

\newcommand{\kth}{^{(k)}}
\newcommand{\kpth}{^{(k+1)}}
\newcommand{\knl}{\mathrm{k}}

\newcommand{\kx}{\mathbf{k}_x}
\newcommand{\kxy}{\mathrm{K}_{xy}}

\newcommand{\bib}{\bi{b}}
\newcommand{\bid}{\bi{d}}
\newcommand{\bix}{\bi{x}}
\newcommand{\biy}{\bi{y}}
\newcommand{\bipsi}{\bi{\psi}}
\newcommand{\bieps}{\bi{\varepsilon}}
\newcommand{\mycdd}{\mycd{\bid}}
\newcommand{\mycdx}{\mycd{\bix}}
\newcommand{\mycdw}{\mycd{\wt}}
\newcommand{\mycda}{\mycd{a}}
\newcommand{\mycdb}{\mycd{\bib}}
\newcommand{\mycdg}{\mycd{g}}
\newcommand{\mycdeta}{\mycd{\eta}}
\newcommand{\mycdp}{\mycd{p}}
\newcommand{\mycdq}{\mycd{\bi{q}}}
\newcommand{\mycdr}{\mycd{r}}
\newcommand{\mycds}{\mycd{s}}
\newcommand{\mycdt}{\mycd{t}}
\newcommand{\mycdu}{\mycd{u}}
\newcommand{\mynew}{^{\mathrm{new}}}
\newcommand{\myold}{^{\mathrm{old}}}
\newcommand{\myprev}{^{[l]}}
\newcommand{\mynext}{^{[l+1]}}
\newcommand{\mycdf}{\mycd{f}}
\newcommand{\gnorm}{_{G_i}^2}
\newcommand{\sgnorm}{_{G_i}}
\newcommand{\inorm}{_{G_i^{-1}}^2}
\newcommand{\jnorm}{_{G_j^{-1}}^2}
\newcommand{\sinorm}{_{G_i^{-1}}}
\title{Maximing the Margin in the Input Space}

\author{
Shotaro Akaho \\
AIST Neuroscience Research Institute\\
1--1 Central 2, Umezono, Tsukuba 3058568 Japan \\
{\texttt{s.akaho@aist.go.jp}}}

\begin{document}

\maketitle

\begin{abstract}
 We propose a novel criterion for support vector machine learning:
 maximizing the margin in the input space, not in the feature (Hilbert) space. 
 This criterion is a discriminative version of the principal curve
 proposed by Hastie et al.
 The criterion is appropriate in particular when the input space is
 already a well-designed feature space with rather small dimensionality.
 The definition of the margin is generalized
 in order to represent prior knowledge.
 The derived algorithm consists of two alternating steps to estimate the
 dual parameters.
 Firstly, the parameters are initialized by the original SVM.
 Then one set of parameters is updated by Newton-like procedure, and
 the other set is updated by solving a quadratic programming problem.
 The algorithm converges in a few steps to a local optimum under mild
 conditions and it preserves the sparsity of support vectors.
 Although the complexity to calculate temporal variables increases
 the complexity to solve the quadratic programming problem for each step
 does not change.
 It is also shown that the original SVM can be seen as a special case.
 We further derive a simplified algorithm which enables us to use
 the existing code for the original SVM.
\end{abstract}

\section{Introduction}
The support vector machine (SVM) is known as one of state-of-the-art
methods especially for pattern recognition
\cite{cortes,mueller,vapnik}.
The original SVM maximizes the margin which is 
defined by the minimum distance between samples 
and a separating hyperplane in a Hilbert space $\mycal H$. 
Even when the dimensionality of $\mycal H$ is very large,
it has been proved that the original SVM has
a bound for a generalization error
which is independent of the dimensionality.
In practice, however, 
the original SVM sometimes gives a very small margin in the input
space, because the metric of the feature space is usually quite different from
that of the input space.
Such a situation is undesirable in particular when the input space
is already a well-designed feature space by using some prior
knowledge\cite{amari,decoste,jaakkola,simard,tsuda}.

This paper gives a learning algorithm to maximize the
margin in the input space.
One difficulty is getting an explicit form of the
margin in the input space, because the classification boundary is curved and
the vertical projection from a sample point to the boundary is not
always unique. We solve this problem by linear approximation
techniques.  The derived algorithm basically consists of iterations
of the alternating two stages as follows:
one is to estimate the projection point and the other is
to solve a quadratic programming to find optimal parameter values.

Such a dual structure appears in other frameworks, such as
EM algorithm and variational Bayes.
Much more related work is the principal curve proposed by
Hastie et al\cite{hastie}. The principal curve finds a curve in a `center'
of the points in the input space.

The derived algorithm is not a gradient-descent type but Newton-like;
hence we have to investigate its convergence property.
It is shown that the derived
algorithm does not always converges to the global optimum, but
it converges to a local optimum under mild conditions.
Some interesting relations to the original SVM are also shown:
the original SVM can be seen as a special case of the algorithm;
and the number of support vectors does not increase so much from the
original SVM.
The algorithm is verified through simple simulations.

\section{Generalized margin in the input space}

We consider a binary classification problem.
The purpose of learning is to construct a map from an $m$-dimensional input
$\bix\in{\Re}^m$ to a corresponding output $y\in\{\pm1\}$ by using
a finite number of samples $(\bix_1,y_1),\ldots,(\bix_n,y_n)$.

Let us consider a linear classifier, 
$y=\mbox{sgn}[f(\bix)]$, where
$f(\bix) \equiv \inner{\wt}{\phi(\bix)} + f_0$; 
$\phi(\bix)$ is a feature of an input $\bix$ in 
a Hilbert space $\mycal H$,
$\wt\in \mycal H$ is a weight parameter
and $f_0\in \Re$ is a bias parameter.
Those parameters $\wt$ and $f_0$ define a separating hyperplane in the
feature space. 
As a feature function $\phi(\bix)$, we only consider a differentiable
nonlinear map.

A margin in the input space is defined by the minimum distance from sample
points to the classification boundary in the input space.
Since the classification boundary forms a complex curved surface,
the distance cannot be obtained in an explicit form, and more
significantly, a projection from a point to the boundary is not unique.

Here, the metric in the input space is not necessary to be Euclidean.
Some Riemannian metric $G(\bix)$ may be defined, which
enables us to represent many kinds of prior knowledge.
For example, the invariance of patterns\cite{mueller,simard} can be implemented
in this form.
Another example is that 
Fisher information matrix is a natural metric,
when the input space is a parameter space
of some probability distribution\cite{amari,jaakkola}.
Although the distance is theoretically preferable to be measured by
the length of a geodesic in the Riemannian space,
it causes computational difficulty.
In our formulation, since we only need a distance from a sample point to
another point, we use a computationally feasible (nonsymmetric) distance
from a sample point $\bix_i$ to another point $\bix$ in the quadratic norm,
\[
\|\bix-\bix_i\|\gnorm =
  (\bix-\bix_i)\transpose G_i(\bix-\bix_i),
\]
where $G_i\equiv G(\bix_i)$.

For simplicity, we mainly consider the hard margin case in which
sample points are separable by a hyperplane in the Hilbert space.
The soft margin case is discussed in the section \ref{sec:soft}.

Let $\myftx_i$ be the closest point on the boundary
surface from a sample point $\bix_i$, and
$\bid_i \equiv \myftx_i - \bix_i$.
Since $\bid_i$ is invariant under a scalar transformation of $(\wt,f_0)$,
we can assume all points are separated with satisfying
\begin{equation}
  \label{eq:constraint}
  \|\bid_i\|\gnorm \ge {1/\inner{\wt}{\wt}},\quad i=1,\cdots,n,
\end{equation}
If we assume at least one of them is an equality,
the margin is given by $1/\sqrt{\inner{\wt}{\wt}}$.
Then we can find the optimal parameter by minimizing
a quadratic objective function $\inner{\wt}{\wt}$
with the constraints (\ref{eq:constraint}) and $y_i f(\bix_i) > 0$.

In order to solve the optimization problem, we start from a solution
of the original SVM and update the solution iteratively.
By two kinds of linearization technique and a kernel trick
which are described in the next section, we obtain
a discriminant function at the $k$-th iteration step in the form of
\begin{equation}
\label{eq:f}
 f(\bi{x})=\sum_{i\in \mathrm{S.V.}} \{a_i\kth \knl(\mycdx_i\kth,\bix) +
  \bib_i\kth{}\transpose \kx(\mycdx_i\kth, \bix)\} + f_0\kth,
\end{equation}
where S.V. is a set of indices of support vectors,
$\knl(\bix,\biy)$ is a kernel function and $\kx(\bix,\biy)$ is its
derivative defined by $\kx(\bix,\biy)\equiv {\partial
\knl(\bix,\biy)/\partial\bix}$.
We have two groups of parameters here: One is of $a_i$, $\bib_i$ and $f_0$
which are parameters of linear coefficients, and the other is
of $\mycdx_i$ which is an estimate of
the projection point $\myftx_i$ and forms base functions.
$a_i$ and $f_0$ are initialized by the corresponding parameters in the 
original SVM and the other parameters are initialized by
$\bib_i=\mathbf0$, $\mycdx_i=\bix_i$.

\section{Iterative QP by linear approximations}
In this section, we overview the derivation of update rules of
those parameters. The resultant algorithm is summarized in sec.\ref{sec:overall}.

\subsection{Linear approximation of the distance to the boundary}
\label{sec:d}
Suppose an estimated projection point $\mycdx_i$ is given,
we can get an approximate distance $\|\bid_i\|\sgnorm$
by a linear approximation\cite{akaho}.
\hfill Taking the Taylor expansion of \\
 $f(\myftx_i)=0$ around $\mycdx_i$
up to the first order,
we obtain a constraint on $\bid_i$,
\[
 f(\mycdx_i) + 
 \nabla f(\mycdx_i)\transpose (\bid_i - \mycdd_i) = 0,
\]
where $\mycdd_i = \mycdx_i-\bix_i$.
Minimizing $\|\bid_i\|\gnorm$ under this constraint,
we have
\begin{equation}
\label{eq:d}
\|\bid_i\|\gnorm = {(\inner{\wt}{\{\phi(\mycdx_i) -
 \bipsi(\mycdx_i)\transpose\mycdd_i \}}+f_0)^2\over 
\|\inner{\wt}{\bipsi(\mycdx_i)}\|\inorm},
\end{equation}
where $\bipsi(\mycdx_i)\equiv
\nabla \phi(\mycdx_i)\in {\mycal H}^m$.
Note that this approximate value is unique, and it is invariant under a
scalar transformation of
$(\wt,f_0)$.
Moreover, the approximation is strictly correct when $\mycdx_i=\myftx_i$
and $\nabla f(\myftx_i)\ne 0$.

\subsection{Linearization of the constraint}
\label{sec:qp}
Using the approximate value of the distance, we have a nonlinear
constraint, 
\begin{equation}
\label{eq:NLconst}
 y_i\left[\inner{\wt}\{\phi(\mycdx_i) -
 \bipsi(\mycdx_i)\transpose\mycdd_i \}+f_0\right]
  \ge {\|\inner{\wt}{\bipsi(\mycdx_i)}\|\sinorm\over\sqrt{\inner{\wt}{\wt}}}.
\end{equation}
Since the constraint is nonlinear for $\wt$, we linearize it around
an approximate solution $\wt=\mycdw$ which is the solution at
a current step.
This linearization not only simplifies the problem, but
also enables us to derive a dual problem.

Let $g_i(\wt)$ be the right hand side of (\ref{eq:NLconst}),
the first order expansion is 
\[
  g_i(\wt) = g_i(\mycdw) +
   \inner{\left({\partial g_i(\mycdw)/\partial\wt}\right)}{(\wt-\mycdw)}.
\]
Now let $\mycdg_i \equiv g_i(\mycdw),
 \mycdeta_i \equiv {\partial g_i(\mycdw)/\partial\wt}$,
then we have a linear constraint for $\wt$,
\begin{equation}
\label{eq:constraint3}
 \inner{\wt}{[y_i\ \{\phi(\mycdx_i) -
\bipsi(\mycdx_i)\transpose\mycdd_i
 \}-\mycdeta_i]}\ge \mycdg_i- f_0 y_i,
\end{equation}
where we used the fact $\inner{\mycdw}{\mycdeta_i}=0$.
Suppose $\mycdq_i \equiv \inner{\mycdw}{\bipsi(\mycdx_i)}$ and
$\mycdr \equiv \inner{\mycdw}{\mycdw}$,
then $\mycdg_i$ and $\mycdeta_i$ are given by
\begin{eqnarray}
\label{eq:h}
 \mycdg_i &=& {1\over \sqrt{\mycdr}}\|\mycdq_i\|\sinorm,\nonumber\\
 \mycdeta_i 
   &=& {1\over \mycdg_i \mycdr} \left\{\mycdq_i\transpose G_i^{-1}
    \bipsi(\mycdx_i) -{1\over\mycdr}\|\mycdq_i\|\inorm\mycdw\right\}.
\end{eqnarray}
By the above linearization, we can derive the dual problem
in a similar way to the original SVM,
\begin{eqnarray}
\lefteqn{W(\bi{\mylag}) = \sum_i \mycdg_i\mylag_i} \nonumber\\
&& -{1\over2}
  \sum_{i,j}\mylag_i\mylag_j [y_i \{\phi(\mycdx_i) -
\bipsi(\mycdx_i)\transpose\mycdd_i
 \}-\mycdeta_i]\cdot[y_j \{\phi(\mycdx_j) -
 \bipsi(\mycdx_j)\transpose \mycdd_j
 \}-\mycdeta_j], \nonumber
\end{eqnarray}
which is maximized under constraints $\mylag_i\ge0$ \\
and $\sum_i\mylag_i y_i = 0$.
The solution $\wt$ is given by 
\begin{equation}
\label{eq:wt}
\wt = \sum_i \mylag_i [y_i \{\phi(\mycdx_i) -
\bipsi(\mycdx_i)\transpose\mycdd_i
 \}-\mycdeta_i].
\end{equation}
Here we can see an apparent relation to the original SVM, i.e.,
by letting $\mycdx_i=\bix_i$, $\mycdeta_i=0$, and $\mycdg_i=1$,
we have the exactly the same optimization problem as the original SVM.

\subsection{Kernel trick}

In order to avoid the calculation of mapping into high dimensional
Hilbert space, SVM applies a kernel trick, by which
an inner product is replaced by a symmetric positive definite
kernel function (Mercer kernel) that is easy to
calculate\cite{ramsey,cortes,mueller,vapnik}.
In our formulation, 
$\inner{\phi(\bix)}{\phi(\biy)}$ is replaced by a Mercer kernel
$\knl(\bix,\biy)$.
We also have to calculate the inner product
related to $\bipsi$ (the derivative of $\phi$).
Let us assume that the kernel function $\knl$ is differentiable.
Then, $\inner{\bipsi(\bix)}{\phi(\biy)}$
is replaced by a vector
$\kx(\bix,\biy)\equiv {\partial \knl(\bix,\biy)/\partial\bix}$,
and $\inner{\bipsi(\bix)}{\bipsi(\biy)\transpose}$
is replaced by a matrix
$\kxy(\bix,\biy)
\equiv {\partial^2 \knl(\bix,\biy)/\partial\bix\partial\biy\transpose}$.

Now we can derive the kernel version of the optimization problem.
In (\ref{eq:wt}), $\mycdeta_i\in \mycal H$ has bases related to
$\bipsi(\mycdx_i)$ and $\mycdw$,
and the solution $\wt$ has bases $\phi(\mycdx_i)$ additionally.
Although $\mycdw$ can have any kinds of bases, we restrict it
in the following form to avoid increasing number of bases.
\[
 \mycdw=\sum_i \{\mycda_i \phi(\mycdx_i) +
  \mycdb_i\transpose \bipsi(\mycdx_i)\}.
\]
Then we have
$\mycdq_i = \sum_j \{ \mycda_j
  \kx(\mycdx_i, \mycdx_j) +
  \kxy(\mycdx_i,\mycdx_j)\mycdb_j
  \}$.
Now let 
\[
 \mycdp_i \equiv \inner{\mycdw}{\phi(\mycdx_i)} =
  \sum_j \{\mycda_j\knl(\mycdx_j,\mycdx_i) + \mycdb_j\transpose
  \kx(\mycdx_j,\mycdx_i)\},
\]
then $\mycdr$ is given by
$\mycdr = \sum_i (\mycda_i \mycdp_i + \mycdb_i\transpose\mycdq_i)$,
and $\mycdg_i$ by (\ref{eq:h}).
Further, let us define additional temporal variables
that represent several terms in the objective function,
\begin{eqnarray*}
 \mycds_{ij} &\equiv& \inner{\{\phi(\mycdx_i) -
\bipsi(\mycdx_i)\transpose\mycdd_i
 \}}{\{\phi(\mycdx_j) -
 \bipsi(\mycdx_j)\transpose \mycdd_j
 \}} \\
 &=& \knl(\mycdx_i,\mycdx_j)+\mycdd_i\transpose
  \kxy(\mycdx_i,\mycdx_j)\mycdd_j
  -\mycdd_i\transpose\kx(\mycdx_i,\mycdx_j)
  -\mycdd_j\transpose\kx(\mycdx_j,\mycdx_i), \\
\mycdt_{ij} &\equiv& \inner{\mycdeta_i}
{\{\phi(\mycdx_j) - \bipsi(\mycdx_j)\transpose\mycdd_j\}}
\\
&=& 
{1\over \mycdg_i \mycdr}\bigg\{\mycdq_i\transpose G_i^{-1}
 \left(\kx(\mycdx_i,\mycdx_j) - \kxy(\mycdx_i,\mycdx_j)\mycdd_j
  \right) 
  - {\|\mycdq_i\|\inorm\over \mycdr}(
  \mycdp_j - \mycdd_j\transpose\mycdq_j)
 \bigg\}, \\
 \mycdu_{ij} &=& \inner{\mycdeta_i}{\mycdeta_j} 
 =
 {1\over \mycdg_i \mycdg_j \mycdr^2}(\mycdq_i\transpose G_i^{-1}\kxy(\mycdx_i,\mycdx_j) G_j^{-1}\mycdq_j 
  -{\|\mycdq_i\|\inorm\|\mycdq_j\|\jnorm\over\mycdr}),
\end{eqnarray*}
then we have the objective function in a kernel form,
\begin{equation}
W(\bi{\mylag}) = \sum_i \mycdg_i\mylag_i
 -{1\over2}\sum_{i,j}\mylag_i\mylag_j (y_i y_j \mycds_{ij} - y_j \mycdt_{ij}-
 y_i \mycdt_{ji}+\mycdu_{ij}),
\label{eq:qp}
\end{equation}
which is maximized under constraints
\begin{equation}
\label{eq:constrainta}
 \mylag_i\ge0, \qquad \sum_i y_i\mylag_i = 0.
\end{equation}

The new parameters can be determined from (\ref{eq:wt}) by
\begin{eqnarray}
\label{eq:newab}
 a_i\kpth &=& \mylag_i y_i + \beta \mycda_i,\nonumber\\
 \bib_i\kpth &=& -\mylag_i\left(y_i\mycdd_i+ {G_i^{-1}\mycdq_i\over
			 \mycdg_i \mycdr}\right) +\beta
 \mycdb_i,
\end{eqnarray}
where
$ \beta = \sum_j{\mylag_j\|\mycdq_j\|\inorm/\mycdg_j\mycdr^2}$.

As for the bias term $f_0$, since the constraint
(\ref{eq:constraint3}) should be satisfied in equality
for $J=\{i\mid\mylag_i\ne0\}$ from
the Kuhn-Tucker condition, we have for any $i\in J$,
\begin{equation}
\label{eq:newf}
 f_0\kpth = y_i \mycdg_i -\sum_j \mylag_j
  (y_j \mycds_{ji} - \mycdt_{ji} - y_i y_j \mycdt_{ij} + y_i \mycdu_{ij})
\end{equation}

From ($\ref{eq:newab}$), we can estimate the number of support vectors.
Let $J_k$ be the indices of nonzero $\mylag_i$'s at the $k$-th step, then
the number of support vectors is bounded from upper by
$|J_0\cup J_1 \cup \cdots \cup J_k|$. Since $J_k$ does not
change much as long as the structure of classification boundary
is similar,
the number of support vectors is expected to be not so larger than
the original SVM.

\subsection{Update of the approximate projection of the points}
To complete the algorithm, we have to consider the update of the approximate value
of the projection point $\mycdx_i$ which is initialized by $\bix_i$, otherwise the convergent solution is not precise
what we want.
If good approximates $\mycdw$ and $\mycdf_0$ of
the solution are given, we can refine $\mycdx_i$
iteratively in the same way as in sec. \ref{sec:d}:
Suppose $\mycdw=\sum_j \{\mycda_j \phi(\mycdx_j\myold) +
\mycdb_j\transpose \bipsi(\mycdx_j\myold)\}$,
the projection point $\mycdx_i$ can be estimated by iterating
the following steps for $l=0,1,2,3,\cdots$,
\begin{equation}
\label{eq:upmycdx}
 \mycdx_i\mynext
   = \bix_i -
   {\mycdq_i\myprev\over\|\mycdq_i\myprev\|\inorm}
   \left[\mycdp_i\myprev
    - (\mycdx_i\myprev{}-\bix_i)\transpose
    \mycdq_i\myprev  + \mycdf_0\right]
\end{equation}
where $\mycdx_i^{[0]}$ is initialized by $\mycdx_i\myold$;
$\mycdp_i\myprev$ and $\mycdq_i\myprev$ are defined in a similar way as
$\mycdp_i$ and $\mycdq_i$,
\begin{eqnarray}
 \mycdp_i\myprev &\equiv& \inner{\mycdw}{\phi(\mycdx_i\myprev)} \nonumber\\
 &=&
  \sum_j \{\mycda_j\knl(\mycdx_j\myold,\mycdx_i\myprev) + \mycdb_j\transpose
  \kx(\mycdx_j\myold,\mycdx_i\myprev)\}, \nonumber \\
 \mycdq_i\myprev &\equiv&
  \inner{\mycdw}{\bipsi(\mycdx_i\myprev)}\nonumber\\
 &=&\sum_j \{ \mycda_j
  \kx(\mycdx_i\myprev, \mycdx_j\myold) +
  \kxy(\mycdx_i\myprev,\mycdx_j\myold)\mycdb_j
  \}.\nonumber
\end{eqnarray}

Note that locally maximum points and saddle
points of the distance are also equilibrium states
of (\ref{eq:upmycdx}). The following proposition guarantees
such a point is not stable.
\begin{proposition}
A point $\mycdx_i\in {\Re}^m$ is an equilibrium state of the
 iteration step (\ref{eq:upmycdx}), when and only when the point
 is a critical point of the distance from $\bix_i$ to the
 separating boundary, i.e.,
 a local minimum, a local maximum or a saddle point.
 The equilibrium state is not stable when the point is a
 local maximum or a saddle point.
\end{proposition}
\textit{Proof:}
It is straightforward to show that a point is
an equillibrium state of the iteration step (\ref{eq:upmycdx}),
only when the point is a critical point of the projection point
$\|\bid_i\|\gnorm$. Without loss of generality,
we can assume the uniform metric case $G_i=I$, because
update rule (\ref{eq:upmycdx}) is invariant of a metric transformation.
We consider the behavior around a critical point $\myftx_i$.
Let $\mycdx_i\myprev=\myftx_i+\bieps$,
for a sufficiently small vector $\bieps$.
One can show that $\mycdx_i\myprev$ is mapped into the separating
hypersurface $f(\bix)=\inner{\mycdw}{\phi(\bix)}+\mycdf_0=0$
for a small $\bieps$ after one step iteration.
Therefore, we only consider the
case $\mycdx_i\myprev$ is on the hypersurface.

Since $\myftx_i$ is a critical point
of the distance, the tangent vector $\nabla f(\myftx_i)$ is
collinear to the distant vector $\bid_i=\myftx_i-\bix_i$, i.e.,
for some constant $\lambda$, it holds
\begin{equation}
 \nabla f(\myftx_i) = \lambda \bid_i.
\end{equation}
Furthermore, if $\mycdx_i\myprev$ is in a point of $f(\bix)=0$,
$\nabla f(\myftx_i)$ is nearly orthogonal to $\bieps$,
i.e.,
\begin{equation}
 \nabla f(\myftx_i)\transpose \bieps \simeq 0.
\end{equation}
By expanding (\ref{eq:upmycdx}) around $\myftx_i$, we have
a new estimation $\mycdx_i\mynext$ by
\begin{equation}
\label{eq:mycdx}
 \mycdx_i\mynext \simeq \myftx_i
 + {1\over\lambda}\nabla^2 f(\myftx_i)\bieps
  - {\bid_i\transpose\nabla^2 f(\myftx_i)\bieps\over\lambda\|\bid_i\|}\bid_i,
\end{equation}
where $\nabla^2 f$ is a hessian matrix of $f(\bix)$.
Without loss of generality, we can take the coordinate of $\bix$ as
follows: the first coordinate is the direction of $\bid_i$, and
the second to the $m$-th coordinates are taken orthogonally such that
an $(m-1)\times(m-1)$ submatrix of $\nabla^2 f(\myftx_i)$
for those coordinates is diagonalized, i.e., $\nabla^2 f(\myftx_i)$
is in the form,
\begin{equation}
 \nabla^2 f(\myftx_i) = \left(
\begin{array}{cccc}
c_1 & & \bi{b}\transpose & \\
 & c_2 & & 0 \\
\bi{b} & & \ddots & \\
 & 0 & & c_m \\
\end{array} \right).
\end{equation}
Under this coordinate system,
since $\varepsilon_1$ is of small order value,
the first element calculated from the second and third term in (\ref{eq:mycdx})
vanishes and we have
\begin{equation}
\mycdx_i\mynext - \myftx_i \simeq {1\over\lambda}
 (0, c_2 \varepsilon_2,\ldots,c_m\varepsilon_m)\transpose.
\end{equation}
The iteration step is stable at $\myftx_i$ only when
$\|\mycdx_i\mynext-\myftx_i\|\le\|\forall\bieps\|$, i.e.,
t$|c_j|< |\lambda|$ for all $j=2,\ldots,m$. \hfill $\Box$

\begin{figure}[tbhp]
  \begin{center}
   \includegraphics[width=.8\textwidth]{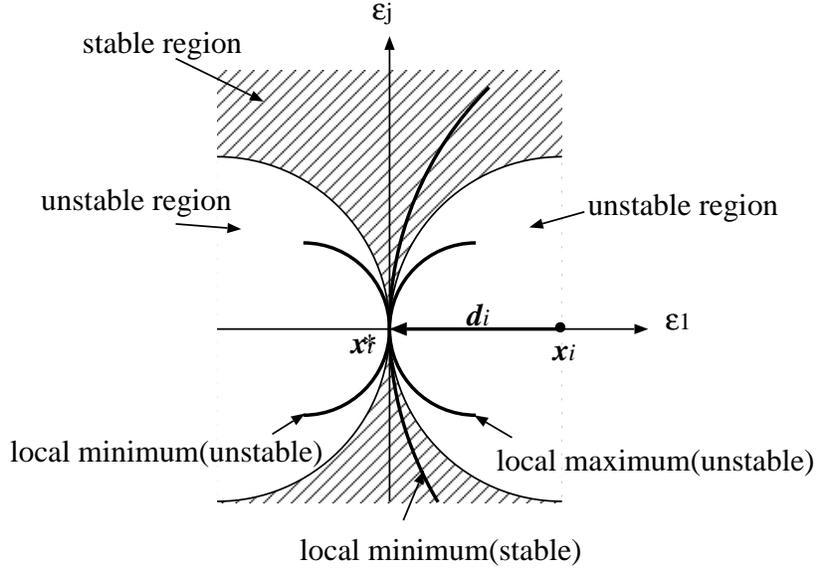}
    \caption{Stability of projection point update}
    \label{fig:stability}
  \end{center}
\end{figure}

The condition for 1-$j$ plane is shown in figure \ref{fig:stability}.

When the point is a local maximum or saddle, the hypersurface is in the unstable
region. However, even in the case of local minimum, there exist an
unstable region, when the hypersurface is stronglly curved.
We can avoid the undesired behavior by slowing down.
For example, first $c_2,\ldots,c_m$ and $\lambda$ are estimated from
$\nabla f$ and $\nabla^2 f$ values at the current estimate,
and then if $c_j < |\lambda|$
for all $j=2,\ldots,m$, the point is to be local minima, then
the movement $\mycdx_i\mynext-\mycdx_i\myprev$
to the axes in which $c_j<-|\lambda|$ should be
shrinked by multiplying some factor $0 < e_j < |\lambda|/|c_j|$.

This computationally intensive
treatment would be usually necessary only
after the several steps, because it is considered
that the unstablity for local minima occurs a small region
relatively to the size of $\bid_i$.

\subsection{Projection of the hyperplane}
\label{sec:proj}
The update of $\mycdx_i$ causes another problem:
We assumed in section \ref{sec:qp}
that $\wt$ and $\mycdw$ have the same bases.
However, $\mycdw$ has bases based on the old $\mycdx_i$, while
we need the new $\wt$ based on the new $\mycdx_i$.
To solve that problem, $\mycdw$ is projected into new bases, i.e.,
from the old one
$\mycdw\myold=\sum_{i\in \mathrm{S.V.}}\{\mycda\myold_i
\phi(\mycdx_i\myold) + \mycdb\myold_i{}\transpose\bipsi(\mycdx_i\myold)
\}$ 
to a new one,
$\mycdw\mynew=\sum_{i\in \mathrm{S.V.}}\{\mycda\mynew_i
\phi(\mycdx_i\mynew) + \mycdb\mynew_i{}\transpose\bipsi(\mycdx_i\mynew)\}$.
Although $\mycdw\mynew$ can have more bases other than S.V.,
we restrict the bases to support vectors to
preserve the sparsity of bases.

There are several possibilities of the projection.
In this paper, we use the one which minimizes the cost function
\begin{equation}
\label{eq:E}
 {1\over2}\sum_{\bix\in T} \{\inner{\mycdw\mynew}{\phi(\bix)} + \mycdf_0\mynew -
  (\inner{\mycdw\myold}{\phi(\bix)} + \mycdf_0\myold)\}^2, 
\end{equation}
where $T$ is a certain set of $\bix$, and we use $T=$ $\{\bix_i$,
$\mycdx_i\myold$, $\mycdx_i\mynew$; $i=1,\cdots,n\}$.

Minimizing (\ref{eq:E}) leads to a simple least square problem, which can
be solved by linear equations.
Another possibility of the cost function is
$\|\mycdw\mynew-\mycdw\myold\|^2$, which leads to another set of
linear equations.

\subsection{Overall algorithm and the convergence property}
\label{sec:overall}

Now let us summarize the algorithm below.
\par
\bigskip
\par
\noindent{\textbf{\strut Algorithm 1: Algorithm to maximize the margin
in the input space}}
\hrule 
\strut Initialization step:
       Let the solution of the original SVM be
       $a_i\zeroth$ and $f_0\zeroth$; 
       let $\bib_i\zeroth=\mathbf0$ and $\mycdx_i\zeroth=\bix_i$.
\par\noindent
For $k=0,1,2,\ldots$, repeat the following steps until convergence:
\begin{enumerate}
 \item Update of $\mycdx_i$:
       Calculate $\mycdx_i\kpth$ by
       applying (\ref{eq:upmycdx}) iteratively to $\mycdx_i\kth$.
 \item Projection of hyperplane:
       Calculate $\mycda_i$, $\mycdb_i$ and $\mycdf_0$ based on
       $\mycdx_i\kpth$ by
       a certain projection method from $a_i\kth$, $\bib_i\kth$ and $f_0\kth$
       based on $\mycdx_i\kth$ (sec.\ref{sec:proj}).
 \item QP step: Solve the QP problem (\ref{eq:qp})
       with respect to $\mylag_i$.
 \item Parameter update:
       Calculate $a_i\kpth$, $\bi{b}_i\kpth$ and $f_0\kpth$ by
       (\ref{eq:newab}) and (\ref{eq:newf}).
\end{enumerate}
The discriminant function at the $k$-th step is given by (\ref{eq:f}).
\par\smallskip
\hrule
\bigskip

Although Algorithm 1 does not always converge to the global minimum,
we can prove the following proposition concerning about the convergence
of the algorithm.
\begin{proposition}
Equilibrium points of Algorithm 1 are critical points of the margin in
 the input space.
The algorithm is stable, when the update rule of $\mycdx_i$ (\ref{eq:upmycdx})
 is stable for all $i$ (see also Proposition 1).
\end{proposition}
This proposition can be proved basically by proposition 1 and the fact that
the linearization of QP is almost exact by a small
perturbation of $\wt$.
As in the case of (\ref{eq:upmycdx}), we can modify the algorithm by
slowing down in (\ref{eq:d}) and (\ref{eq:upmycdx}) so that
the equilibrium state is stable when and only when the margin
is locally optimal.
However, we don't use it in the simulation because the case
that the local minimum is unstable is expected to be rare.

Another problem of Algorithm 1 is that each iteration step does not
always increase the margin monotonically.  
Although it is usually faster than gradient type algorithms,
the algorithm sometimes does not improve the solution of the original
SVM at all.
Because the original SVM can be seen as a special case of the algorithm,
we can use some annealing technique, for example, updating temporal
variables and parameters more gradually from their initial values.
However, for simplicity, we use a crude method in the simulation
as follows: Repeat several 
steps of the algorithm (5 steps in the simulation) and then choose
the best solution which gives the largest estimated value of the margin.

As for the complexity of the algorithm, we need $O(m^2 n^2)$ space
and $O(m^3 n^2)$ time complexity to calculate temporal variables
if the computation of a kernel function is $O(m)$,
while the original SVM requires $O(n^2)$ space and $O(m n^2)$ time.
Those calculation can be pararellized easily.
This complexity is not so different when $m$ is comparatively small.
Once the variables are calculated, the complexity for QP is just the same.
Therefore, as far as the calculation for temporal variables
is comparative to the QP time,
the proposed algorithm is comparative to the original SVM.
If the Algorithm 1 is heavy because of the large $m$, we can use
a simplified algorithm as shown in the section \ref{sec:simple}.

As for the iteration of QP which is carried out usually for a few steps,
since a current solution is an estimate of the solution,
it may be able to reduce the complexity
of the QP at the next iteration step.

\section{Simulation results}
\label{sec:simulation}

\begin{figure}[tbhp]
 \includegraphics[width=.8\textwidth]{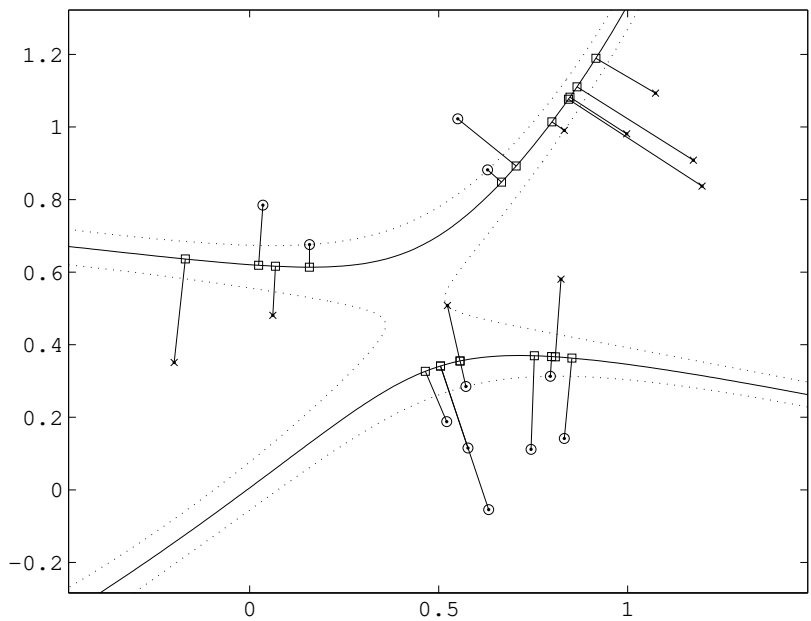}
\caption{Result of the original SVM (margin .040).
Circles ($\circ$) and crosses ($\times$) are positive and negative
samples. Squares ($\Box$) represent estimates of the projection
of the points by applying (\ref{eq:upmycdx}) for 10 steps.}
\label{fig:svm}
\end{figure}

\begin{figure}[tbhp]
 \includegraphics[width=.8\textwidth]{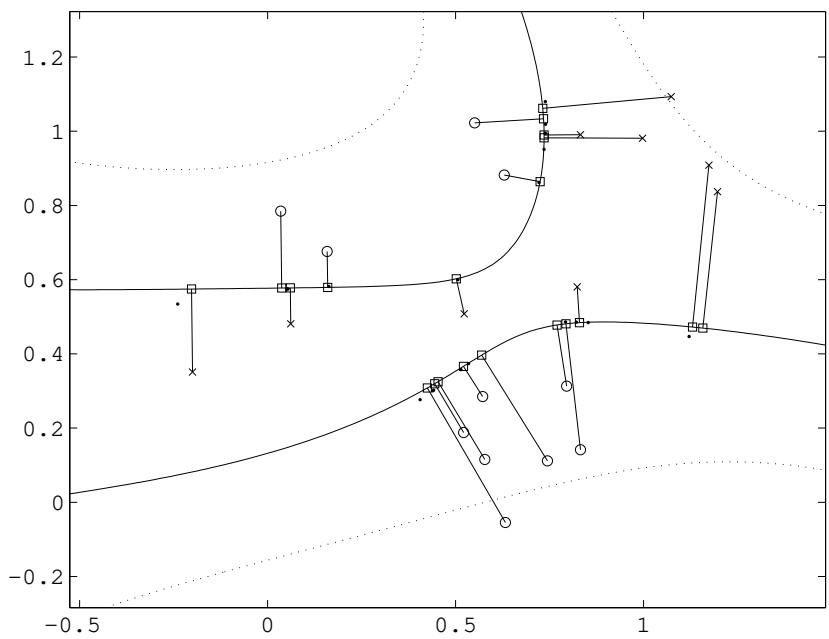}
\caption{Result of the algorithm 1 (after 5 steps, margin .096)
 for the same data set as fig.\ref{fig:svm}}
\label{fig:alg1}
\par\bigskip
\end{figure}

In this section, we give a simulation result for 
artificial data sets in order to verify the proposed algorithm
 and to examine the basic performance.
20 training samples and 1000 test samples are randomly drawn from
positive and negative distribution, each of which is a
Gaussian mixture of 3 components with
uniformly distributed centers $[0,1)^2$ and
fixed spherical variance $\sigma^2=0.2^2$.
The kernel function used here is a spherical Gaussian kernel with
$\sigma^2=1^2$.
The metric is taken to be Euclidean (i.e., $G_i$ is the unit matrix).
Figure \ref{fig:svm} and \ref{fig:alg1}
show an example of results by the original SVM
(initial condition) and the proposed algorithm (after 5 steps).
In this case, the margin value increases from 0.040 to 0.096.
Such a simulation is repeated for 100 sets of samples with different random
numbers.

\begin{figure}[tbhp]
 \includegraphics[width=.8\textwidth]{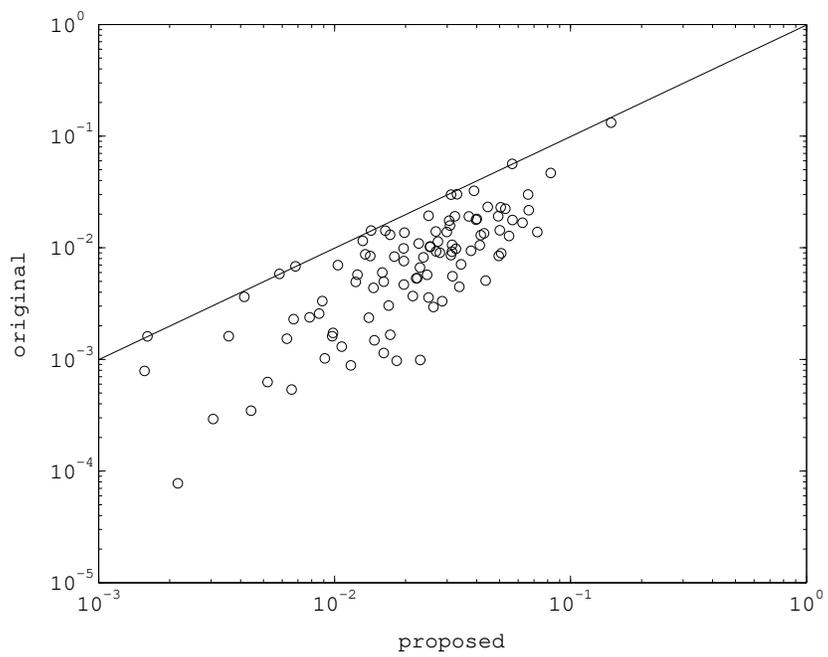}
\caption{Margin comparison with the original SVM for 100 runs
 (log-log scale)}
\label{fig:margin}
\end{figure}

The estimated margins
in the input space for the original and proposed
algorithm is shown in figure \ref{fig:margin} (log-log scale).
By the crude algorithm described in the
previous section, there are 4 cases among 100 runs that cannot improve the
margin of the original SVM. The ratios of the margin are distributed
from 1.00 (no improvement) to 27.9.

\begin{figure}[tbhp]
 \includegraphics[width=.8\textwidth]{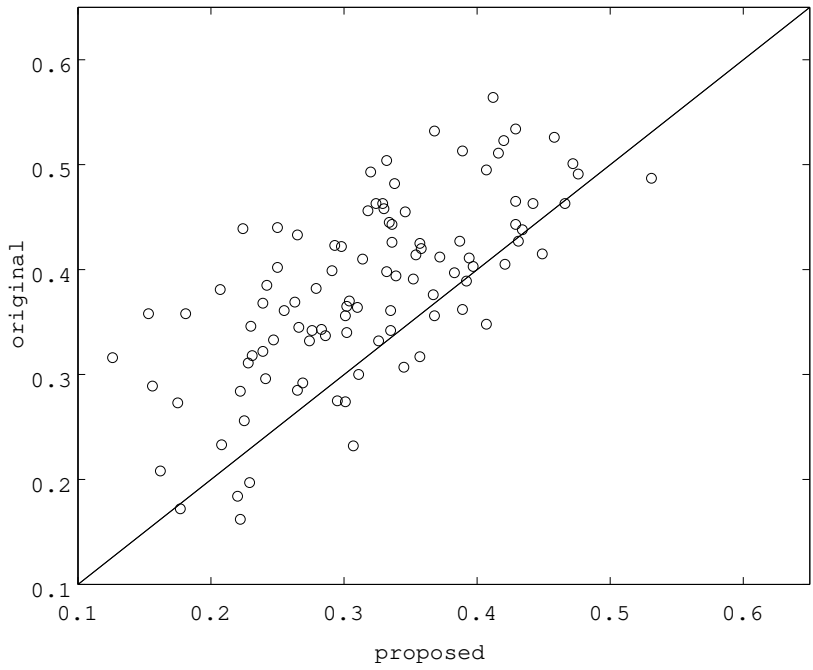}
\caption{Test error comparison with the original SVM for 100 runs}
\label{fig:error}
 \par\bigskip
\end{figure}

The misclassification errors
for test samples is shown in figure \ref{fig:error}.
The ratios of error distributed between [0.40(best),1.37(worst)].

This results indicates that the margin in the input space
is efficient to improve the generalization performance in average, but
there are cases that cannot reduce the generalization error
even when the margin in the input space increases.

\section{Soft margin}
\label{sec:soft}

For noisy situation, the hard margin classifier often overfits
samples. 
There are several possibitilities to incorporate the soft margin,
here we give a simple one.
The soft margin can be derived by introducing slack variables $z_i$
into the optimization problem.
If we use a soft constraint in the form
\begin{equation}
\label{eq:constraint5}
 \inner{\wt}{[y_i\ \{\phi(\mycdx_i) -
\bipsi(\mycdx_i)\transpose\mycdd_i
 \}-\mycdeta_i]}\ge \mycdg_i-f_0 y_i - z_i,
\end{equation}
and adding penalty for the slack variables,
\begin{equation}
 {1\over2}\inner{\wt}{\wt} + C\sum_i z_i,
\end{equation}

By this modification, only the constraint (\ref{eq:constrainta}) for
$\mylag_i$ is changed to
\begin{equation}
 0\le\mylag_i\le C, \qquad \sum_i y_i\mylag_i = 0,
\end{equation}
which is the same constraint as the soft margin of the original SVM.
However, the geometrical meaning of (\ref{eq:constraint5}) in the space
is not clear. It is a future work to introduce a natural soft constraint
in the input space.

\section{Simplified algorithm for a high dimensional case}

\label{sec:simple}

Although Algorithm 1 achieves the precise solution, the computation
costs is high for large dimensionality of inputs.
In this section, we give a simplified algorithm.

If we don't update $\mycdx_i$, the first and the second steps of Algorithm 1
is not necessary any more. This simplification makes Algorithm 1
a little simpler because all $\mycdd_i$ terms vanish.
However, let us consider further simplification.

We have shown the relation to the original SVM:
the original SVM can be derived $\mycdg_i=1$ and $\mycdeta_i=0$.
Since $\mycdeta_i$ causes many temporal variables,
we only maintain $\mycdg_i$.
Then all the terms related to $\mycdb_i$'s vanish.

Consequently,
the above simplifications lead to the algorithm much like the original
SVM. In fact, the existing code for the original SVM can be used as follows:

For each step, first $\mycdg_i$ is calculated,
\begin{equation}
 \mycdg_i = {\|\sum_j a_i\kx(\bix_i,\bix_j)\|\sinorm\over
  \sqrt{\sum_{j,k}a_j\kth a_k\kth \knl(\bix_j,\bix_k)}}.
\end{equation}
Then, by letting the $(i,j)$ element of kernel matrix be
$\knl(\bix_i,\bix_j) / \mycdg_i\mycdg_j$, the original SVM for this
kernel matrix gives the solution for each step of the simplified algorithm.

\section{Conclusion}
We have proposed a new learning algorithm to find a kernel-based
classifier that maximizes the margin in the input space.
The derived algorithm consists of an alternating optimization between
the foot of perpendicular and the linear coefficient parameters.
Such a dual structure appears in other frameworks, such as
EM algorithm, variational Bayes, and principal curve.

There are many issues to be studied about the algorithm, for example,
analyzing the generalization performance theoretically and
finding an efficient algorithm that reduces the complexity and
converges more stably.
It is also an interesting issue to extend our framework to other
problems than classification, such as regression\cite{akaho,otsu,mueller}.

In this paper, we have assumed that the kernel function is given and fixed.
Recently, several techniques and criteria to choose a kernel function
have been proposed extensively. We expect that
those techniques and much other knowledge for the original SVM
can be incorporated in our framework.
Applying the algorithm to real world data is also important.


\begin{thebibliography}{12}
 \bibitem{akaho} S. Akaho, Curve fitting that minimizes the mean square of 
perpendicular distances from sample points, {\it SPIE Vision Geometry
	 II} (also found in {\it Selected SPIE Papers on CD-ROM}, 
	 8, 1999), 237--244 (1993)

 \bibitem{amari}
 S. Amari, {\it Differential Geometrical Methods in
	 Statistics}, Springer-Verlag (1984)

 \bibitem{cortes}
 C. Cortes and V.N. Vapnik, Support vector machines,
	 {\it Machine Learning}, 20, pp. 273--297 (1995)

 \bibitem{decoste}
 D. DeCoste and B. Sch\"olkopf, Training invariant
	 support vector machines, {\it Machine Learning}, 46(1),
	 pp. 161--190 (2002)

 \bibitem{hastie}
	 T. Hastie and W. Stuetzle, Principal curves,
	 {\it Journal of the American Statistical Association}, 84(406),
	 pp. 502--516 (1989)

 \bibitem{jaakkola}
 T.S. Jaakkola and D. Haussler, Exploiting generative
	 models in discriminative classifiers, {\it NIPS 11},
	 pp. 487--493 (1998)

 \bibitem{mueller}
 K.R. M\"uller, S. Mika, G. R\"atch, K. Tsuda,
	 B.Sch\"olkopf, An Introduction to Kernel-Based Learning
	 Algorithms, {\it IEEE Trans. on Neural Networks}, 12,
	 pp. 181--201 (2001)

 \bibitem{otsu}
 N. Otsu, Karhunen-Loeve line fitting and a linearly
	 measure. In {\it IEEE Proc. of ICPR'84}, pp. 486--489 (1984)

 \bibitem{ramsey}
 J.O. Ramsey, B.W. Silverman, {\it Functional Data Analysis},
	 Springer-Verlag (1997)
	 
 \bibitem{simard}
  P.Y. Simard, Y.A. Le Cun, J.S. Denker, B. Victorri,
	 Transformation Invariance in Pattern Recognition -- Tangent
	 Distance and Tangent Propagation, in {\it Neural Networks:
	 Tricks of the Trade}, G. Orr and K.-R. M\"uller, eds.,
	 Springer-Verlag, vol.1524, pp.239--274 (1998)

 \bibitem{tsuda}
  K. Tsuda, M. Kawanabe, G. R\"atsch, S. Sonnenburg, K.R. M\"uller,
         A New Discriminative Kernel from Probabilistic Models,
	 {\it NIPS 14} (2001)

 \bibitem{vapnik}
  V.N. Vapnik, {\it The Nature of Statistical
	 Learning Theory}, Springer-Verlag (1995)
\end{thebibliography}
\end{document}